# Land Cover Semantic Segmentation Using ResUNet

Vasilis Pollatos*, Loukas Kouvaras**, Eleni Charou***

*NTUA, Athens, Greece, vaspoll97@gmail.com

**Harokopio University, Athens, Greece

*** NCSR Demokrtios, Athens, Greece  exarou@iit.demokritos.gr

**ABSTRACT** In this paper we present our work on developing an automated system for land cover classification. This system takes a multiband satellite image of an area as input and outputs the land cover map of the area at the same resolution as the input. For this purpose convolutional machine learning models were trained in the task of predicting the land cover semantic segmentation of satellite images. This is a case of supervised learning. The land cover label data were taken from the CORINE Land Cover inventory and the satellite images were taken from the Copernicus hub. As for the model, U-Net architecture variations were applied. Our area of interest are the Ionian islands (Greece). We created a dataset from scratch covering this particular area. In addition, transfer learning from the BigEarthNet dataset [1] was performed. In [1] simple classification of satellite images into the classes of CLC is performed but not segmentation as we do. However, their models have been trained into a dataset much bigger than ours, so we applied transfer learning using their pretrained models as the first part of out network, utilizing the ability these networks have developed to extract useful features from the satellite images (we transferred a pretrained ResNet50 into a U-Res-Net). Apart from transfer learning other techniques were applied in order to overcome the limitations set by the small size of our area of interest. We used data augmentation (cutting images into overlapping patches, applying random transformations such as rotations and flips) and cross validation. The results are tested on the 3 CLC class hierarchy levels and a comparative study is made on the results of different approaches.

**INTRODUCTION** Modern AI technologies, such as deep learning, can be utilized in various fields of natural science to automate and underpin procedures traditionally carried out by humans. Remote sensing nowadays provides a great amount of data of high quality which are updated on a daily basis. Another important thing is that these data are easily produced and are open to the public in contrast to other sources, such as aerial photography that are of higher quality but are more expensively and less massively produced. For some problems (in our case land cover recognition) the resolution of the open remote sensing data (10m for sentinel-2) is adequate. The big data of remote sensing can be fed into machine learning models to develop automated systems that analyse this data and carry out useful tasks. Labeled data are the most useful ones, as they can be utilised for the purposes of supervised learning that solves a great range of problems.

CLC provides a huge labeled dataset. It contains maps for the most part of Europe for the last three decades. Our goal is to train models to predict the labels of the CLC dataset. Most research done in this field is about assigning one or more land cover labels into a whole satellite image patch (which can take an area of several square kilometres). Our approach to the problem is more general, trying to construct a semantic segmentation of the satellite image into the full range of the land cover classes provided by the Corine Land Cover inventor, at the maximal resolution provided by sentinel-2 satellite images, which is 10m. The classes of CLC are hierarchical. We are testing the ability of the models to predict the classes on each one of the hierarchical levels. As expected, we see that the superclasses on the higher levels are discriminated with greater accuracy than the subclasses on the lower levels.

Corine Land Cover has a wide variety of applications, underpinning various Community policies in the domains of environment, but also agriculture, transport, spatial planning. Developing a system that automates the production of CLC maps to some extent is important because CLC needs to be updated every few years. Creating these maps is a burdensome and time-consuming job for the human and even so the accuracy of the produced maps isn't perfect. An automatic land cover classification system could help develop such maps in the future, track down sudden or short term changes that happen to the land cover (for example due to natural disasters or due to fast track rural and urban development). It could also be applied to areas that are not included in the CLC.

State of the art deep learning models were used and the training and testing were done in the area of Ionio. This is a case of work on a relatively small area with special geological and natural features. It is also an area of varying morphology and landscapes and small scale land cover characteristics that can hardly be detected on the resolution provided by sent-2 images. Similar approaches can be used for training and testing in other areas covered by the sentinel-2 satellites. As a first step we trained a simple U-Net from scratch in the area of interest. Recently, a similar research was done in the TU Berlin, developing the BigEarthNet. They perform simple classification of satellite images into the classes of CLC but not segmentation as we do. However, their models have been trained into a dataset much bigger than ours, so we applied transfer learning using their pretrained models as the first part of out network, utilizing the ability these networks have developed to extract useful features from the satellite images (we transferred a pretrained ResNet50 into a U-Res-Net). Apart from transfer learning other techniques were applied in order to overcome the limitations set by the small size of our area of interest. We used data augmentation (cutting images into overlapping patches, applying random transformations such as rotations and flips) and cross validation.

**RELATED WORK** Land Cover Recognition gathers a lot of interest in the research community. In our work we apply transfer learning from the models trained in BigEarthNet [1]. The BigEarthNet dataset contains 590,326 non-overlapping image patches of size 1200m ×1200m distributed over 10 european countries (Austria, Belgium, Finland, Ireland, Kosovo, Lithuania, Luxembourg, Portugal, Serbia, Switzerland). Each image patch is annotated by multiple land-cover classes (i.e., multi-labels) that are provided from the CORINE Land Cover database of the year 2018 (CLC 2018). They train models that take each patch as input and predict the classes appearing in this patch. They solve a simpler problem than ours, because the resolution of the output of their models is 1200m, while the resolution of our predicted maps is 10m. However, their models have been trained on a dataset much bigger than ours and have learned to extract useful features from the images (encoding) that are later on decoded to solve their task. We are using the pretrained encoder of a res-net-50 trained on BigEarthNet as the encoder part of a unet-like architecture to solve our semantic segmentation problem. This approach has also been adopted by [3]. UNet architecture was introduced in [21]. ResNetUnet, the architecture we are using, is commonly used for such problems. In [5] a sophisticated ResNetUnet that performs multitasking achieved state of the art results for the ISPRS 2DPotsdam dataset. One of the subproblems solved in this multitasking is finding the class boundaries, which is also proposed in [10]. However, as far as our problem is concerned, these methods are applied on high resolution images of urban areas and may be of little use for our problem. In order to conquer the limitations set by our small dataset, data augmentation is applied as in [12], [13], [22]. In our work we used Sentinel-2 bands with 10m resolution and bands with 20m resolution. Others have used multisource data including optical data and Sentinel-1 radar measurements [14], [15], [16]. Multi-temporal data viewing the same area on different timestamps is another approach taken in [16],[17],[18],[19]. In order to deal with missing labels active learning [19],[20], self-learning [18] and weakly supervised learning [6], [7], [8] is performed.

## METHODOLOGY

**I) Dataset** Our dataset was created by multispectral satellite images of the Ionian Islands downloaded from Copernicus for the period of 2018 and part of the CLC 2018 that covers the Ionian Islands. CLC vector files were georeferenced together with the Copernicus images, turned into raster with 10m resolution and altogether were clipped in the same bounds creating tiffs for each one of the islands. These tiffs were cut into patches of size 1,28x1,28 km (128x128 pixels) with some high degree of overlap. X_data consists of these patches having the satellite image bands as features for each pixel and Y_data consists of the corresponding CLC patches. Our networks are trained to solve the task of predicting the CLC label for each pixel of the input patch, given the band measurements for each pixel of the input patch. So we are trying to find a function f such that Y_data = f (X_data). This is a case of supervised learning.

Our area of interest is distributed over 6 Ionian islands (Corfu, Paxi, Lefkada, Kalamos, Kefalonia, Zante) and the coast of Parga.

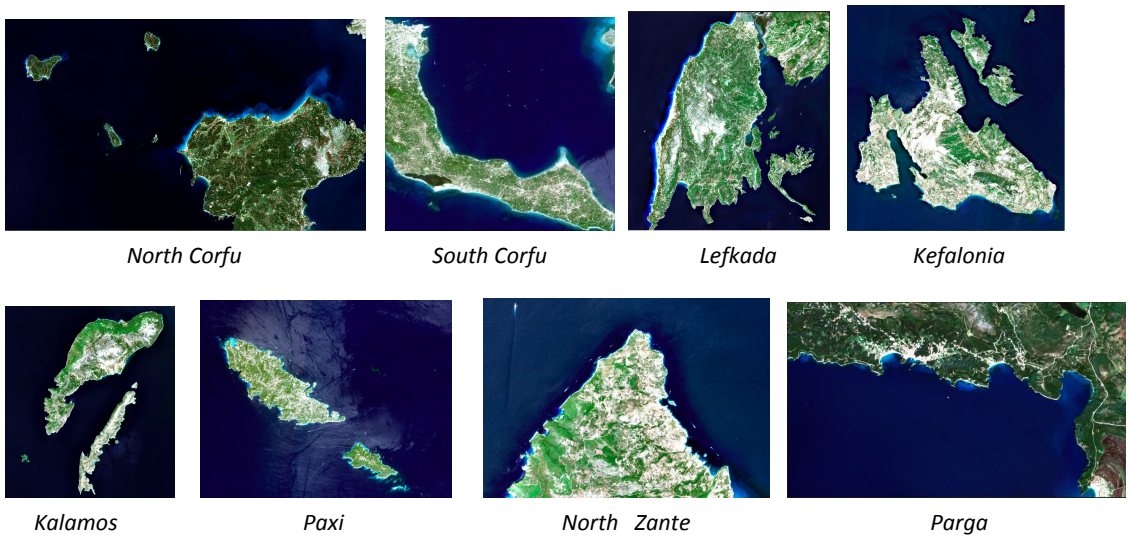

*North Corfu*   *South Corfu*   *Lefkada*   *Kefalonia*

*Kalamos*   *Paxi*   *North Zante*   *Parga*

**Figure 1.** Satellite images on the area of interest

For each area we have the sentinel-2 10m resolution bands (R,G,B, infrared), the sentinel-2 20m resolution bands (b05, b06, b07, b8A, b11 and b12) and the corine land cover class label for each pixel. In our problem the satellite image bands are the inputs to our network and the clc classes the expected output.

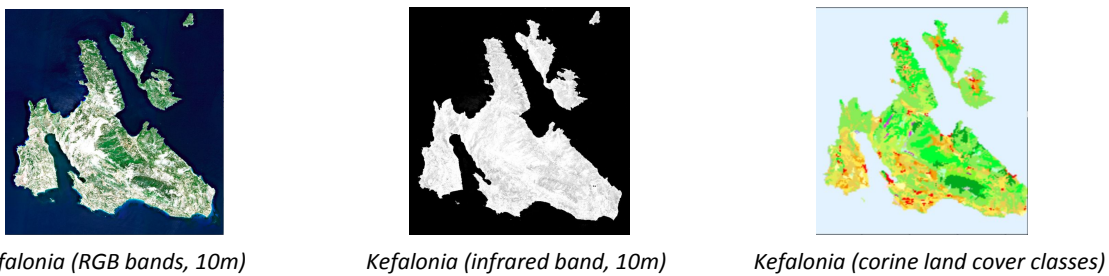

*Kefalonia (RGB bands, 10m)*   *Kefalonia (infrared band, 10m)*   *Kefalonia (corine land cover classes)*

**Figure 2.** Structure of our dataset

Corine Land Cover classes are hierarchical into three levels. Our approach is training the models on the full range of the corine land cover classes and then testing them on each level separately.

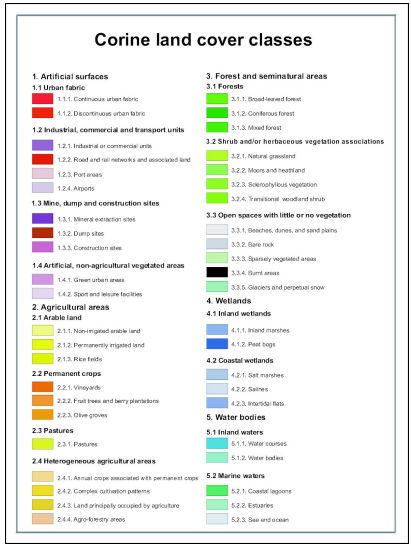

Out of the total of 45 clc classes the following 32 classes where found in the area of interest:

111, 112,
121, 123, 124,
131,
141, 142

211, 212,
221, 222, 223,
231,
242, 243,

311, 312, 313,
321, 323, 324,
331, 332, 333,

411,
421, 422,

512,
521, 522, 523

**Figure 3.** CLC color legend

The area of interest has to be splitted into training and test sets. Due to the small size of our dataset we chose not to use a validation set for the fine tuning of hyperparameters such as the number of epochs. The training process was stopped when the loss function started to converge and not when it was minimal for the validation set. We are performing cross validation so the area of interest has to be divided into a number of subsets of approximately same size . The area of interest was partitioned into the following 6 subsets:
1. north Corfu,    2. south Corfu,    3. west Kefalonia,    4. east Kefalonia,    5. Lefkada,    6. Paxi+North Zante+Kalamos+Parga
The splitting into training and validation sets is done 6 times, so that each time a different subset is the validation set and the remaining 5 are the training set.
Each area is cut into overlapping patches. The overlaps are a form of data augmentation. Patch size is 1.28 km x 1.28 km and the hop between adjacent patches is 0.64 km in each direction (longitude and latitude). Two memory optimisations were applied. Firstly, patches are stored by defining only their limits in the original satellite image and  the cutting is only performed on dataloading. Secondly, patches containing only sea are  discarded ( e.g. the blue square in the right image below). This is a good practice because it turns out that the models are able to learn to recognise the sea almost perfectly even without those patches. It also reduces class imbalancement, as sea patches are the most frequent ones in our area.

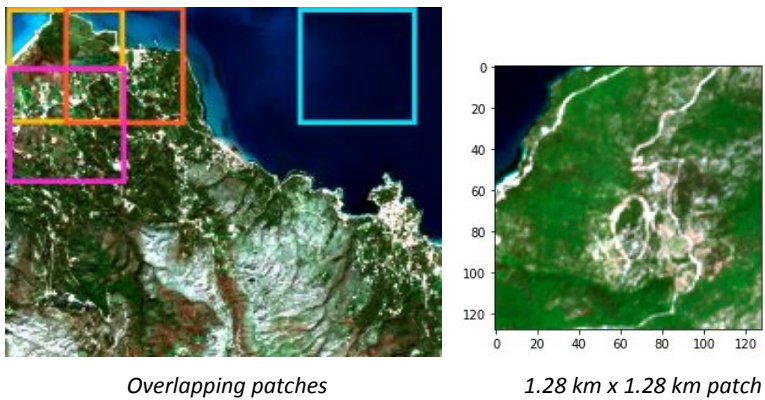

*Overlapping patches*          *1.28 km x 1.28 km patch*

**Figure 4.**  Cutting the original satellite images into patches.

For the transfer learning experiments data needed to be standardised using the same mean and std values as the base model. On dataloading random flips and rotations were applied for the purposes of data augmentation.

**II) Models** Two different approaches were followed. The first approach was to train a baseline UNet from scratch into the area of interest. The second approach was to perform transfer learning. The transfer learning UNet model has a ResNet-50 architecture on the encoder part and the weights of the encoder are initialised to the values of the weights of a ResNet-50 trained on the BigEarthNet . The figure below shows the exact architecture of the transfer learning model. There are approximately 66.000.000 trainable parameters on this model. A more complex version of this model that applied no compression on the outputs of the encoder that were passed to the decoder through shortcuts had  91.000.000 trainable parameters and improved fitting on the training set but didn't seem to generalise better than the model presented below.

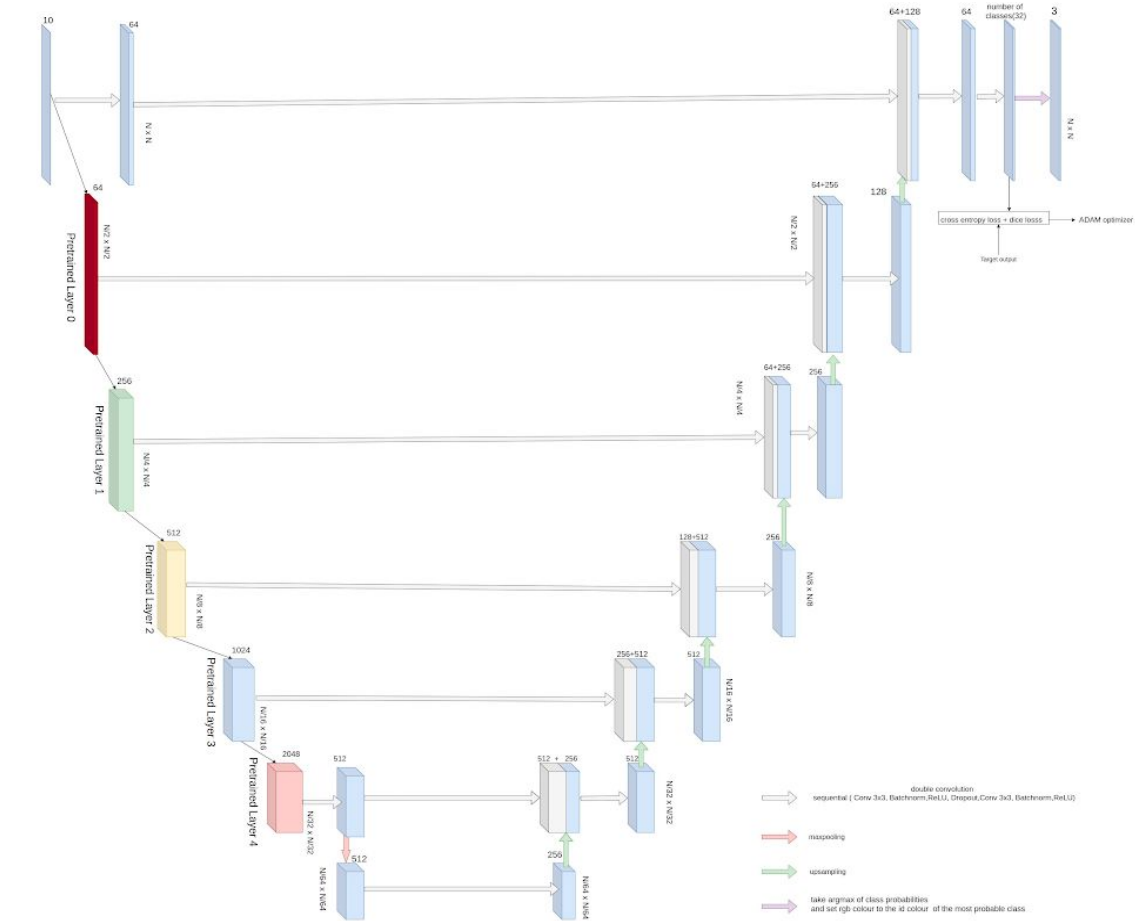
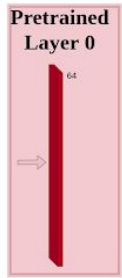
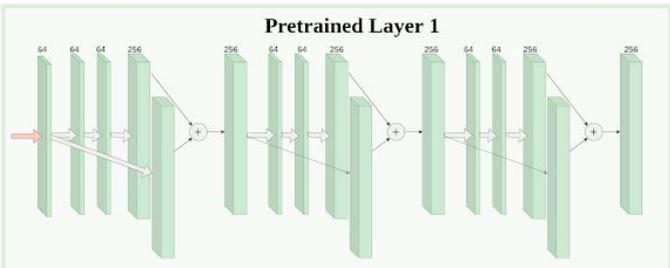
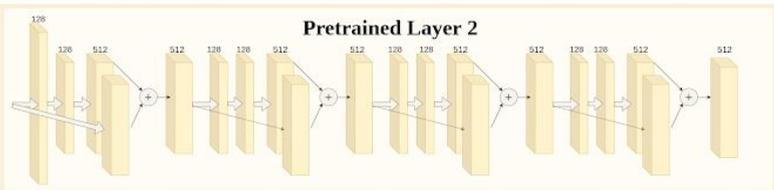
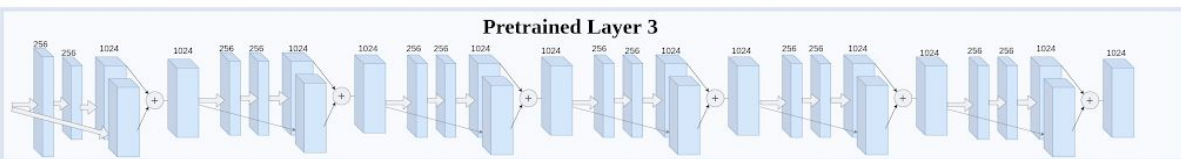
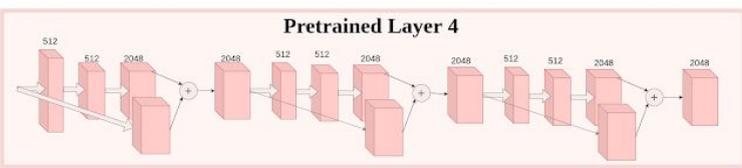

**Figure 5.** Architecture of the transfer learning model

The baseline UNet model that was trained from scratch solved an easier problem, as the output and the ground truth land cover images had a resolution of 100m.

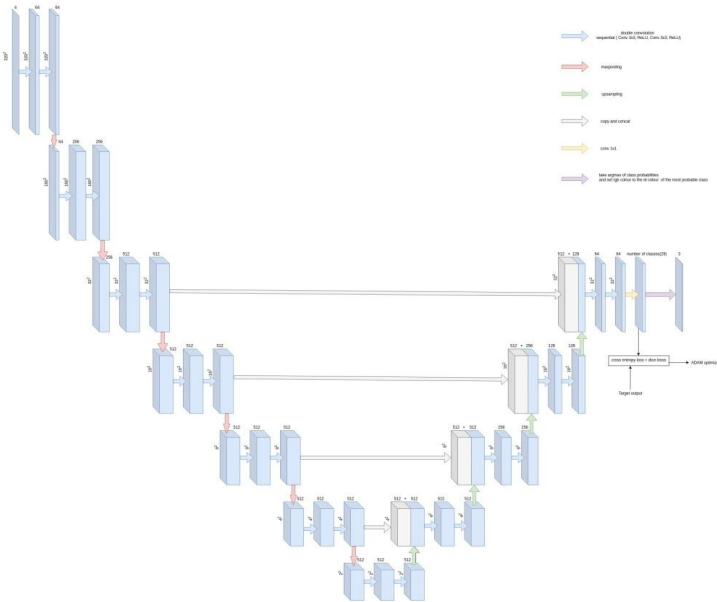

**Figure 6.** Architecture of the baseline UNet model

**III) Training** We are trying to solve a semantic segmentation problem and a composite dice and a binary cross entropy loss with logits criterion is used. The two loss criteria are summed, each one with a weight factor of 0.5. We experimented with positive weights $p_c$ in the bce:
$l_c(y,t)=L_c = \{l_{1,c}, \ldots, l_{N,c}\}$, $l_{n,c} = -w_{n,c}[p_c t_{n,c} * \log(\text{sigmoid}(y_{n,c})) + (1-t_{n,c}) * \log(1 - \text{sigmoid}(y_{n,c}))]$, where c is the class number.
Setting $p_c$ = (number of negative samples of class c)/(number of positive samples of class c) for class balancing deteriorated our results.
Adam optimiser is used to achieve fitting in the training data. initial LR=$5*10^{-4}$ and it gradually decreases with the use of a scheduler. The complexity of our model requires the use of regularization techniques. We applied dropout, with rate 0-0.2 for the outer layers and 0.3-0.4 for the inner hidden layers.
For the first epochs of the training, the weights of the base transfer learning model remain frozen. We unfreeze them when the learning process starts to converge, dropping at the same time the learning rate. As we can see below, unfreezing the base model on epoch 80 causes some instability. However, after some epochs the loss returns to the low values it had before the unfreezing. The pretrained encoder seems to work properly without further training, but the unfreezing brings some slight improvements so we perform it.
Training was executed on google colab.

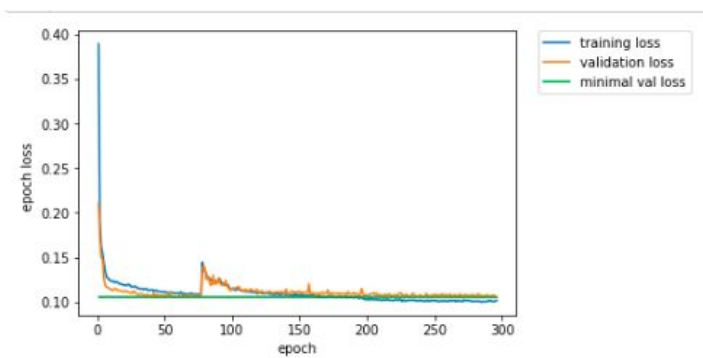

**Figure 7.** Learning curve

**Experiments** Several versions of the problem are being examined. Firstly, training from scratch was done on the area of interest. A baseline model shown in figure 6 was used. The produced maps had a resolution 100m. The visual results and the metrics for the validation set are presented below:

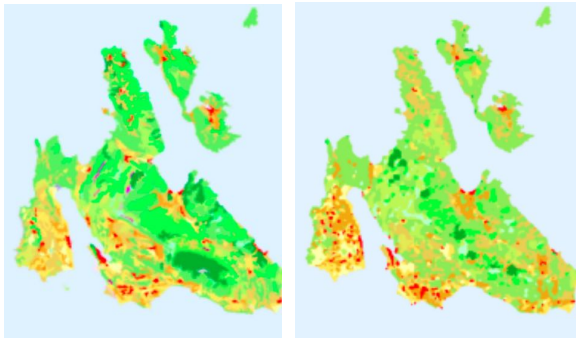

*Kefalonia (target)     Kefalonia (prediction)*

**Figure 8.** Validation results for the model that was trained from scratch

The pixel level classification metrics below are measured on the third level of the clc class hierarchy for the classes that were found on the validation set.

accuracy =  0.787          f1_macro =  0.160          f1_micro =  0.787          f1_weighted =  0.771

precision score =  [1.   0.198 1.   1.   0.   1.   1.   0.0086 1.   0.2435   0.32  0.347 0.321 0.727 0.435 0.479 0.   0.   1.   0.0259 1.   0.   0.992 ]

recall score =  [0.   0.304 0.   0.   0.   0.   0.   0.0013 0.   0.1885 0.   0.194 0.4066 0.4337 0.1948 0.665 0.708 0.   1.   0.   0.0025 0.   1.   0.997 ]

For the transfer learning model a more systematic testing was performed. As mentioned above, 6-fold cross validation was applied. Metrics and maps are calculated for each one of the validation folds. The metrics are taken with respect to each one of the 3 clc class hierarchical levels separately over all pixels (pixel level classification).

The results for each one of the 6 folds are presented below. For the first three we give the validation scores and the map visualisation and for the other three just the visual result, for brevity reasons.

**Fold 1:**

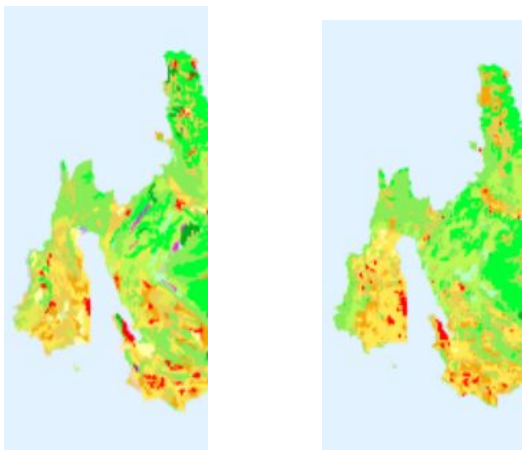

*Target                    Prediction*

**Figure 9.** Validation results for the transfer learning model, west Kefalonia

CORINE CLASS LEVEL 1 :

| class | 1. Artificial Surfaces | 2. Agricultural areas | 3. Forest and seminatural areas | 4. Wetlands | 5. Water bodies |
|---|---|---|---|---|---|
| support | 175093 | 1922887 | 2332193 | 5416 | 8933755 |
| precision | 0.52 | 0.7729 | 0.8075 | 1 | 0.9974 |
| recall | 0.3001 | 0.747 | 0.8537 | 0.0001846 | 0.9985 |

accuracy =  0.92753  
f1_macro =  0.67921  
f1_micro =  0.92753  
f1weighted =  0.92662

CORINE CLASS LEVEL 2 :

| class | support | precision | recall |
|---|---|---|---|
| 1.1 Urban fabric | 119963 | 0.5082 | 0.4122 |
| 1.2 Industrial, commercial and transport units | 40513 | 1 | 2.468e-05 |
| 1.3 Mine, dump and construction sites | 2591 | 1 | 0.0003858 |
| 1.4 Artificial, non-agricultural vegetated areas | 12026 | 0.0002674 | 8.315e-05 |
| 2.1 Arable land | 129947 | 0.134 | 0.02698 |
| 2.2 Permanent crops | 217427 | 0.123 | 0.186 |
| 2.3 Pastures | 234101 | 0.4023 | 0.081 |
| 2.4 Heterogeneous agricultural areas | 1341412 | 0.5679 | 0.6165 |
| 3.1 Forest | 590332 | 0.6192 | 0.5724 |
| 3.2 Shrub and/or herbaceous vegetation associations | 1654002 | 0.6504 | 0.7213 |
| 3.3 Open spaces with little or no vegetation | 87859 | 0.2044 | 0.1991 |
| 4.1 Inland wetlands | 5416 | 1 | 0.0001846 |
| 5.2 Marine waters | 8933755 | 0.9974 | 0.9985 |

accuracy =  0.85329  
f1_macro =  0.4124  
f1_micro =  0.85329  
f1weighted =  0.8522

CORINE CLASS LEVEL 3 :

| class | support | precision | recall |
| --- | --- | --- | --- |
| 1.1.1 Continuous urban fabric | 3443 | 1 | 0.0002904 |
| 1.1.2 Discontinuous urban fabric | 116520 | 0.4734 | 0.3953 |
| 1.2.1 Industrial or commercial units | 28859 | 1 | 3.465e-05 |
| 1.2.3 Port areas | 2606 | 1 | 0.0003836 |
| 1.2.4 Airports | 9048 | 1 | 0.0001105 |
| 1.3.1 Mineral extraction sites | 2591 | 1 | 0.0003858 |
| 1.4.2 Sport and leisure facilities | 12026 | 0.0002674 | 8.315e-05 |
| 2.1.1 Non-irrigated arable land | 129947 | 0.134 | 0.02698 |
| 2.2.1 Vineyards | 18644 | 1 | 5.363e-05 |
| 2.2.3 Olive groves | 198783 | 0.123 | 0.2034 |
| 2.3.1 Pastures | 234101 | 0.4023 | 0.081 |
| 2.4.2 Complex cultivation patterns | 536157 | 0.3528 | 0.4831 |
| 2.4.3 Land principally occupied by agriculture, with significant areas of natural vegetation | 805255 | 0.4042 | 0.3624 |
| 3.1.2 Coniferous forest | 74557 | 0.002304 | 1.341e-05 |
| 3.1.3 Mixed forest | 515775 | 0.5519 | 0.5836 |
| 3.2.1 Natural grassland | 401405 | 0.4262 | 0.5664 |
| 3.2.3 Sclerophyllous vegetation | 1248545 | 0.5923 | 0.6059 |
| 3.2.4 Transitional woodland/shrub | 4052 | 4.215e-05 | 0.0002467 |
| 3.3.2 Bare rock | 27305 | 1 | 3.662e-05 |
| 3.3.3 Sparsely vegetated areas | 60554 | 0.1394 | 0.1944 |
| 4.1.1 Inland marshes | 5416 | 1 | 0.0001846 |
| 5.2.3 Sea and ocean | 8933755 | 0.9974 | 0.9984 |

accuracy = 0.81346  
f1_macro = 0.37624  
f1_micro = 0.81346  
f1weighted = 0.81504

**Fold 2:**

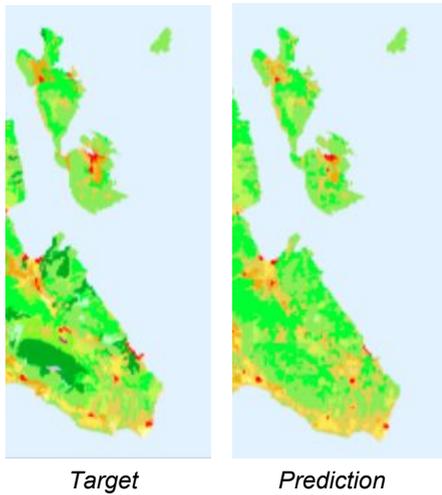

*Target        Prediction*

**Figure 10.** Validation results for the transfer learning model, east Kefalonia

CORINE CLASS LEVEL 1 :

| class | 1. Artificial Surfaces | 2. Agricultural areas | 3. Forest and seminatural areas | 5. Water bodies |
|---|---|---|---|---|
| support | 70026 | 1004333 | 3066378 | 9949503 |
| precision | 0.823 | 0.7099 | 0.9494 | 0.9975 |
| recall | 0.4614 | 0.8556 | 0.8895 | 0.9993 |

accuracy =  0.96251
f1_macro =  0.83431
f1_micro =  0.96251
f1weighted =  0.96409

CORINE CLASS LEVEL 2 :

| class | support | precision | recall |
|---|---|---|---|
| 1.1 Urban fabric | 56640 | 0.8264 | 0.5547 |
| 1.3 Mine, dump and construction sites | 2552 | 1 | 0.0003917 |
| 1.4 Artificial, non-agricultural vegetated areas | 10834 | 0.3934 | 0.04513 |
| 2.1 Arable land | 21436 | 0.6883 | 0.5296 |
| 2.2 Permanent crops | 209584 | 0.6226 | 0.654 |
| 2.3 Pastures | 37121 | 0.3445 | 0.6391 |
| 2.4 Heterogeneous agricultural areas | 736192 | 0.611 | 0.7511 |
| 3.1 Forest | 1116452 | 0.7122 | 0.7187 |
| 3.2 Shrub and/or herbaceous vegetation associations | 1878253 | 0.7847 | 0.7261 |
| 3.3 Open spaces with little or no vegetation | 71673 | 0.8346 | 0.1002 |
| 5.2 Marine waters | 9949503 | 0.9975 | 0.9993 |

accuracy =  0.91362
f1_macro =  0.6004
f1_micro =  0.91362
f1weighted =  0.91511

CORINE CLASS LEVEL 3 :

| class | support | precision | recall |
| --- | --- | --- | --- |
| 1.1.2 Discontinuous urban fabric | 56640 | 0.8264 | 0.5547 |
| 1.3.1 Mineral extraction sites | 2552 | 1 | 0.0003917 |
| 1.4.2 Sport and leisure facilities | 10834 | 0.3934 | 0.04513 |
| 2.1.1 Non-irrigated arable land | 21436 | 0.6883 | 0.5296 |
| 2.2.3 Olive groves | 209584 | 0.6226 | 0.654 |
| 2.3.1 Pastures | 37121 | 0.3445 | 0.6391 |
| 2.4.2 Complex cultivation patterns | 210391 | 0.704 | 0.552 |
| 2.4.3 Land principally occupied by agriculture, with significant areas of natural vegetation | 525801 | 0.4826 | 0.6791 |
| 3.1.2 Coniferous forest | 419030 | 1 | 2.386e-06 |
| 3.1.3 Mixed forest | 697422 | 0.5146 | 0.8313 |
| 3.2.1 Natural grassland | 252838 | 0.7181 | 0.5809 |
| 3.2.3 Sclerophyllous vegetation | 1401252 | 0.6837 | 0.7481 |
| 3.2.4 Transitional woodland/shrub | 224163 | 1 | 4.461e-06 |
| 3.3.2 Bare rock | 13180 | 1 | 7.587e-05 |
| 3.3.3 Sparsely vegetated areas | 58493 | 0.86 | 0.1228 |
| 5.2.3 Sea and ocean | 9949503 | 0.9975 | 0.9993 |

accuracy =  0.88019
f1_macro =  0.559
f1_micro =  0.88019
f1weighted =  0.89214

**Fold 3:**

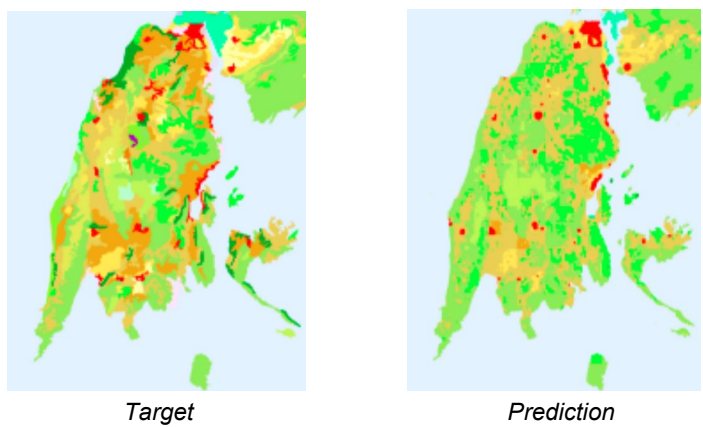

*Target*          *Prediction*

**Figure 11.** Validation results for the transfer learning model, Lefkada
Metrics for the validation set:

CORINE CLASS LEVEL 1 :

| class | 1. Artificial Surfaces | 2. Agricultural areas | 3. Forest and seminatural areas | 4. Wetlands | 5. Water bodies |
|---|---|---|---|---|---|
| support | 176535 | 1869077 | 1861405 | 12270 | 5354057 |
| precision | 0.7811 | 0.8004 | 0.7135 | 0.005529 | 0.9907 |
| recall | 0.3617 | 0.6425 | 0.8799 | 0.0006519 | 0.9982 |

accuracy = 0.88929
f1_macro = 0.61471
f1_micro = 0.88929
f1weighted = 0.89035

CORINE CLASS LEVEL 2 :

| class | support | precision | recall |
|---|---|---|---|
| 1.1 Urban fabric | 108422 | 0.6693 | 0.5016 |
| 1.2 Industrial, commercial and transport units | 3147 | 1 | 0.0003177 |
| 1.3 Mine, dump and construction sites | 4267 | 1 | 0.0002343 |
| 1.4 Artificial, non-agricultural vegetated areas | 60699 | 0.002079 | 1.647e-05 |
| 2.1 Arable land | 108101 | 0.2607 | 0.003996 |
| 2.2 Permanent crops | 552365 | 0.7296 | 0.0475 |
| 2.3 Pastures | 166819 | 0.001908 | 5.994e-06 |
| 2.4 Heterogeneous agricultural areas | 1041792 | 0.4505 | 0.6323 |
| 3.1 Forest | 393277 | 0.2725 | 0.4209 |
| 3.2 Shrub and/or herbaceous vegetation associations | 1417396 | 0.6113 | 0.7278 |
| 3.3 Open spaces with little or no vegetation | 50732 | 0.2842 | 0.002089 |
| 4.2 Coastal wetlands | 12270 | 0.005533 | 0.0006519 |
| 5.2 Marine waters | 5354057 | 0.9907 | 0.9982 |

accuracy =  0.78517
f1_macro =  0.37775
f1_micro =  0.78517
f1weighted =  0.78474

CORINE CLASS LEVEL 3 :

| class | support | precision | recall |
|---|---|---|---|
| 1.1.2 Discontinuous urban fabric | 108422 | 0.6693 | 0.5016 |
| 1.2.3 Port areas | 3147 | 1 | 0.0003177 |
| 1.3.1 Mineral extraction sites | 4267 | 1 | 0.0002343 |
| 1.4.2 Sport and leisure facilities | 60699 | 0.002079 | 1.647e-05 |
| 2.1.1 Non-irrigated arable land | 108101 | 0.2607 | 0.003996 |
| 2.2.1 Vineyards | 7541 | 1 | 0.0001326 |
| 2.2.3 Olive groves | 544824 | 0.7296 | 0.04818 |
| 2.3.1 Pastures | 166819 | 0.001908 | 5.994e-06 |
| 2.4.2 Complex cultivation patterns | 244647 | 0.4023 | 0.2524 |
| 2.4.3 Land principally occupied by agriculture, with significant areas of natural vegetation | 797145 | 0.3345 | 0.5491 |
| 3.1.1 Broad-leaved forest | 20546 | 1 | 4.867e-05 |
| 3.1.2 Coniferous forest | 137230 | 0.000408 | 7.287e-06 |
| 3.1.3 Mixed forest | 235501 | 0.1906 | 0.4898 |
| 3.2.1 Natural grassland | 150340 | 0.49 | 0.5003 |
| 3.2.3 Sclerophyllous vegetation | 1197070 | 0.5248 | 0.6725 |
| 3.2.4 Transitional woodland/shrub | 69986 | 1 | 1.429e-05 |
| 3.3.1 Beaches, dunes, sands | 12323 | 0.2842 | 0.008601 |
| 3.3.3 Sparsely vegetated areas | 38409 | 1 | 2.603e-05 |
| 4.2.1 Salt marshes | 2929 | 0.003458 | 0.001706 |
| 4.2.2 Salines | 9341 | 1 | 0.000107 |
| 5.2.1 Coastal lagoons | 85666 | 0.7482 | 0.2571 |
| 5.2.3 Sea and ocean | 5268391 | 0.98 | 0.998 |

accuracy =  0.73935
f1_macro =  0.32759
f1_micro =  0.73935
f1weighted =  0.74674

**Fold 4:**

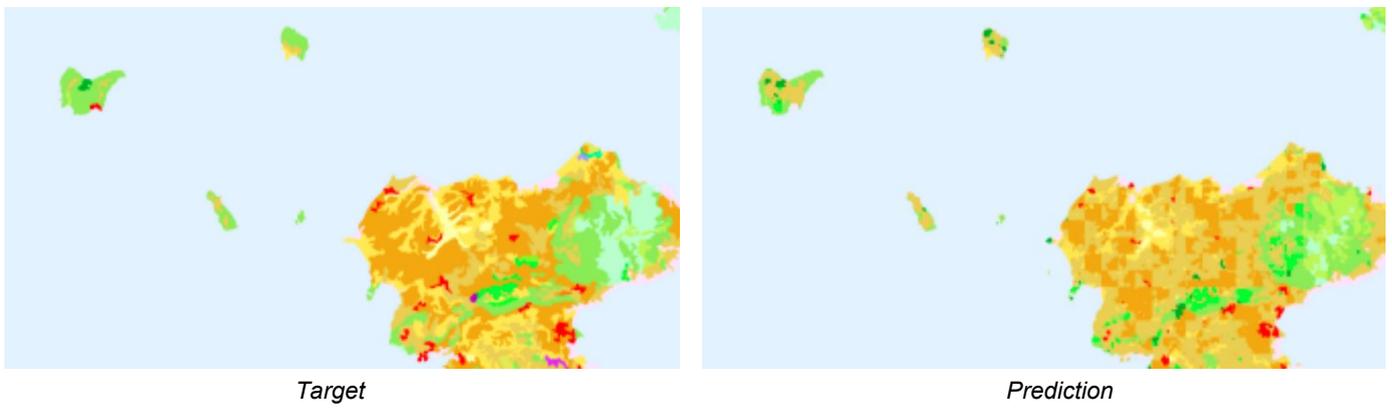

*Target*  *Prediction*

**Figure 12.** Validation results for the transfer learning model, North Corfu

**Fold 5:**

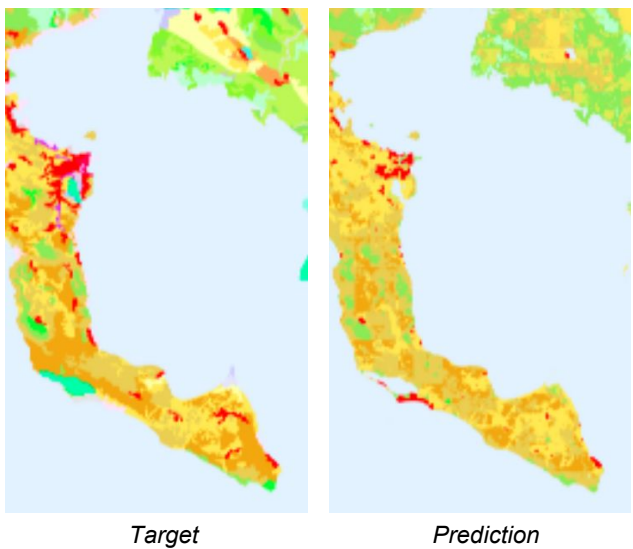

*Target*  *Prediction*

**Figure 13.** Validation results for the transfer learning model, South Corfu

**Fold 6:**

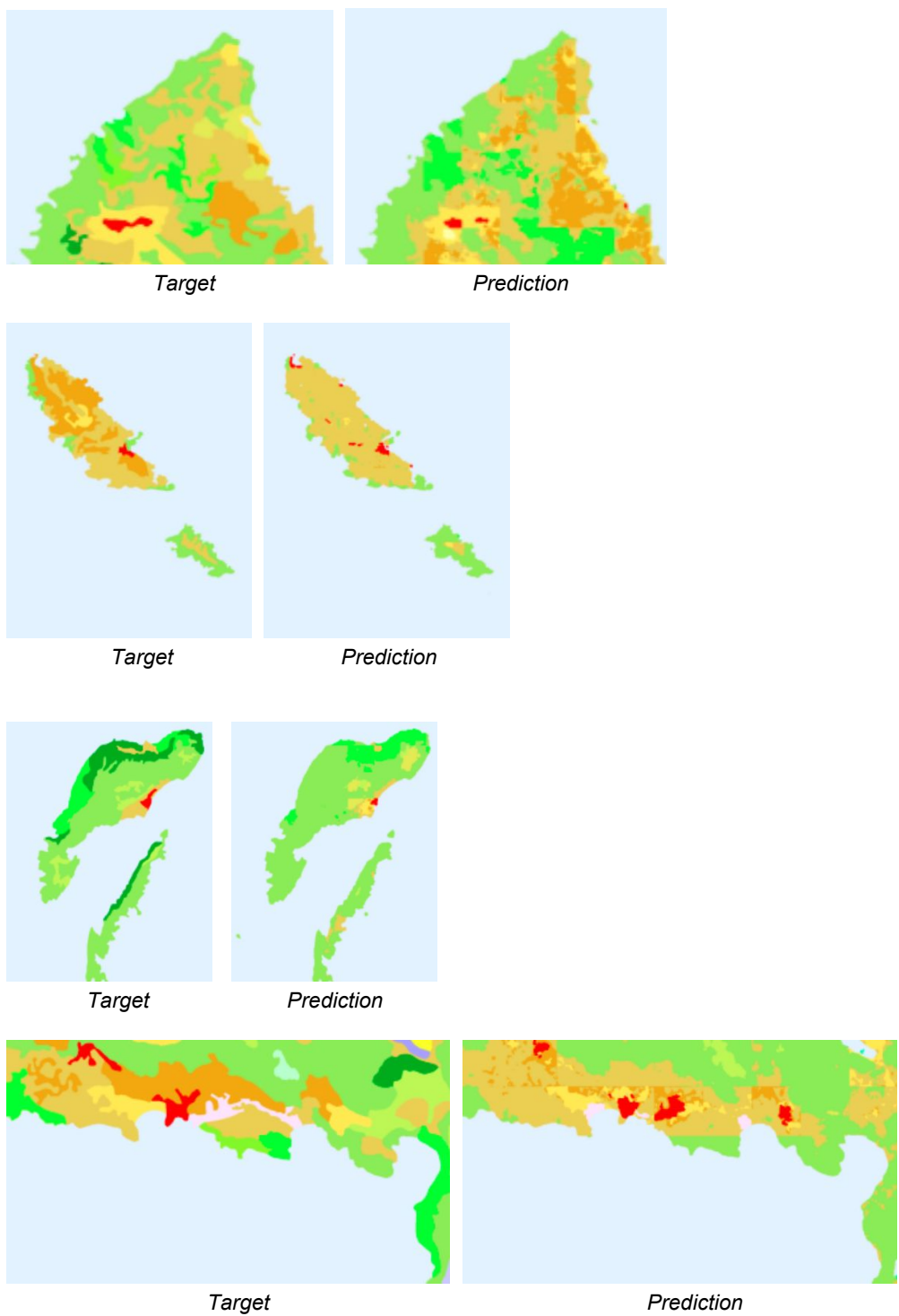

**Figure 14.** Validation results for the transfer learning model (north Zante, Paxoi, Kalamos, Parga)

In the experiments presented above our method was to keep a continuous area as a validation set, for example a whole island. Now we present a different approach where the validation patches are randomly distributed over the area of interest. This is also a realistic problem, where the experts sparsely assign land cover labels on the area of interest and the remaining unlabeled areas are predicted by a model trained on the neighbouring labeled ones. To make sure that the training and the validation set have no common elements we skipped data augmentation via overlaps, but the flips and rotations are still used. We split the area of interest into train and validation with a ratio of 70, 30 respectively.

The metrics for the validation are presented below.

CORINE CLASS LEVEL 1 :

| class | 1. Artificial Surfaces | 2. Agricultural areas | 3. Forest and seminatural areas | 4. Wetlands | 5. Water bodies |
|---|---|---|---|---|---|
| support | 113717 | 1349974 | 1291539 | 7743 | 743203 |
| precision | 0.8639 | 0.8457 | 0.8078 | 1 | 0.9744 |
| recall | 0.06163 | 0.8162 | 0.8998 | 0.0001291 | 0.9916 |

accuracy =  0.85792
f1_macro =  0.68526
f1_micro =  0.85792
f1weighted =  0.85891

CORINE CLASS LEVEL 2 :

| class | support | precision | recall |
|---|---|---|---|
| 1.1 Urban fabric | 77380 | 0.855 | 0.08965 |
| 1.2 Industrial, commercial and transport units | 4145 | 1 | 0.0002412 |
| 1.3 Mine, dump and construction sites | 1825 | 1 | 0.0005476 |
| 1.4 Artificial, non-agricultural vegetated areas | 30367 | 1 | 3.293e-05 |
| 2.1 Arable land | 55779 | 1 | 0.00285 |
| 2.2 Permanent crops | 445942 | 0.523 | 0.6887 |
| 2.3 Pastures | 53487 | 1 | 1.87e-05 |
| 2.4 Heterogeneous agricultural areas | 794766 | 0.5563 | 0.5009 |
| 3.1 Forest | 270234 | 0.6143 | 0.5493 |
| 3.2 Shrub and/or herbaceous vegetation associations | 949394 | 0.65 | 0.8134 |
| 3.3 Open spaces with little or no vegetation | 71911 | 0.5789 | 0.07298 |
| 4.1 Inland wetlands | 4506 | 1 | 0.0002219 |
| 4.2 Coastal wetlands | 3237 | 1 | 0.0003088 |
| 5.2 Marine waters | 743203 | 0.9744 | 0.9916 |

accuracy =  0.67743
f1_macro =  0.40289
f1_micro =  0.67743
f1weighted =  0.68708

CORINE CLASS LEVEL 3 :

| class | support | precision | recall |
| --- | --- | --- | --- |
| 1.1.1 Continuous urban fabric | 7396 | 1 | 0.0001352 |
| 1.1.2 Discontinuous urban fabric | 69984 | 0.855 | 0.09912 |
| 1.2.1 Industrial or commercial units | 2855 | 1 | 0.0003501 |
| 1.2.3 Port areas | 669 | 1 | 0.001493 |
| 1.2.4 Airports | 621 | 1 | 0.001608 |
| 1.3.1 Mineral extraction sites | 1825 | 1 | 0.0005476 |
| 1.4.1 Green urban areas | 207 | 1 | 0.004808 |
| 1.4.2 Sport and leisure facilities | 30160 | 1 | 3.316e-05 |
| 2.1.1 Non-irrigated arable land | 55779 | 1 | 0.00285 |
| 2.2.1 Vineyards | 10122 | 1 | 9.878e-05 |
| 2.2.2 Fruit trees and berry plantations | 9040 | 1 | 0.0001106 |
| 2.2.3 Olive groves | 426780 | 0.5224 | 0.7187 |
| 2.3.1 Pastures | 53487 | 1 | 1.87e-05 |
| 2.4.2 Complex cultivation patterns | 267127 | 0.3909 | 0.5091 |
| 2.4.3 Land principally occupied by agriculture, with significant areas of natural vegetation | 527639 | 0.471 | 0.3283 |
| 3.1.1 Broad-leaved forest | 12029 | 1 | 8.313e-05 |
| 3.1.2 Coniferous forest | 76451 | 1 | 0.09269 |
| 3.1.3 Mixed forest | 181754 | 0.4287 | 0.5532 |
| 3.2.1 Natural grassland | 174214 | 0.4769 | 0.741 |
| 3.2.3 Sclerophyllous vegetation | 701163 | 0.5691 | 0.7445 |
| 3.2.4 Transitional woodland/shrub | 74017 | 1 | 1.351e-05 |
| 3.3.1 Beaches, dunes, sands | 7253 | 1 | 0.0001379 |
| 3.3.2 Bare rock | 7519 | 1 | 0.000133 |
| 3.3.3 Sparsely vegetated areas | 57139 | 0.5789 | 0.09184 |
| 4.1.1 Inland marshes | 4506 | 1 | 0.0002219 |
| 4.2.1 Salt marshes | 1788 | 1 | 0.000559 |
| 4.2.2 Salines | 1449 | 1 | 0.0006897 |
| 5.2.1 Coastal lagoons | 25642 | 1 | 3.9e-05 |
| 5.2.3 Sea and ocean | 717561 | 0.9416 | 0.9925 |

accuracy = 0.59871
f1_macro = 0.28225
f1_micro = 0.59871
f1weighted = 0.62438

Finally we are going to present some examples that show the performance of our model. All the predictions presented are on validation data. The number on the top of each image on the left indicates the fold number (6-fold cross validation).

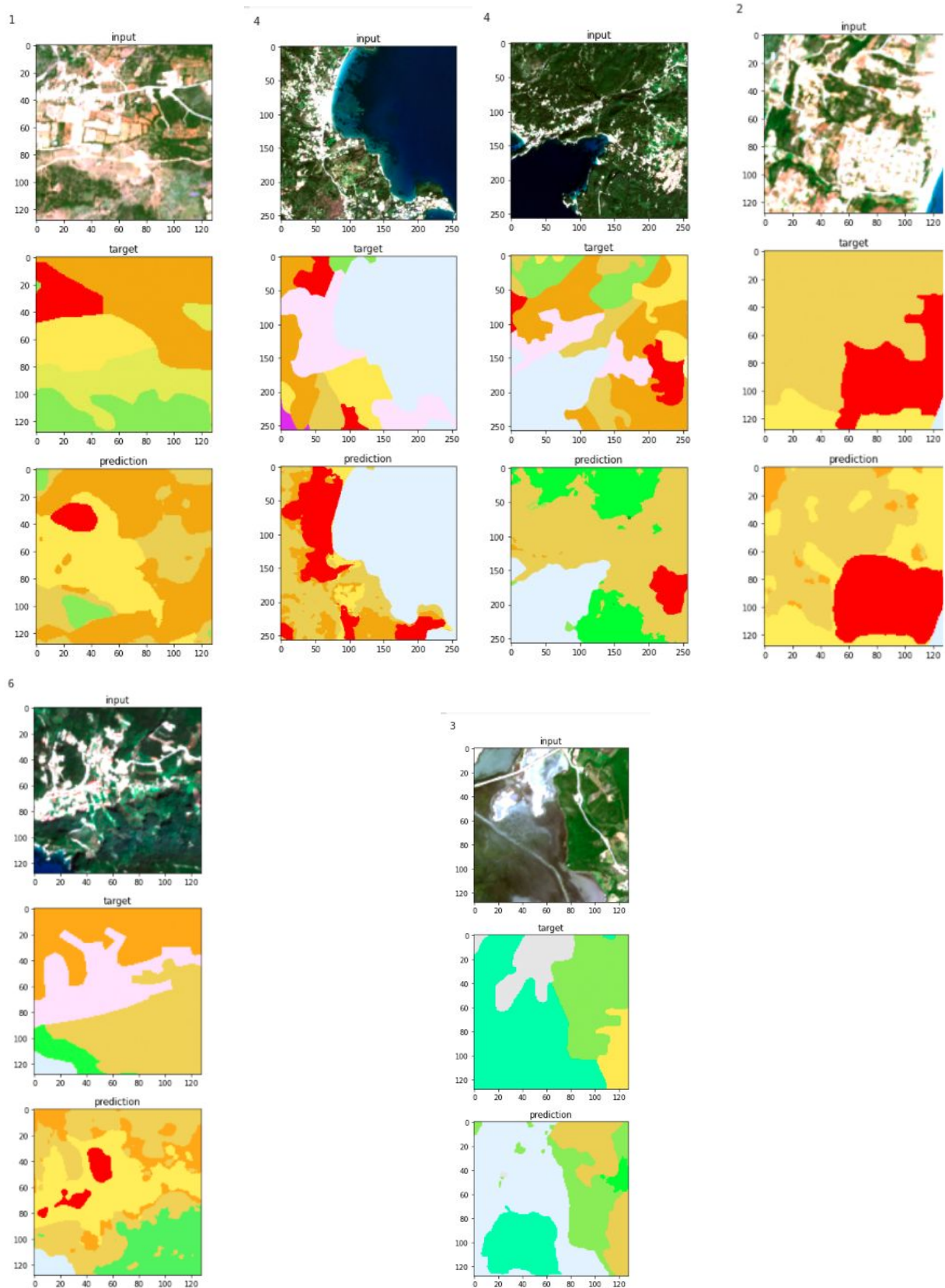

**Figure 15.** Finding uncommon classes (villages and marshes)

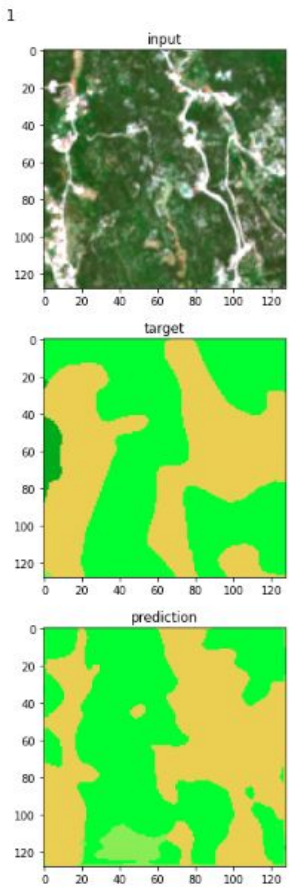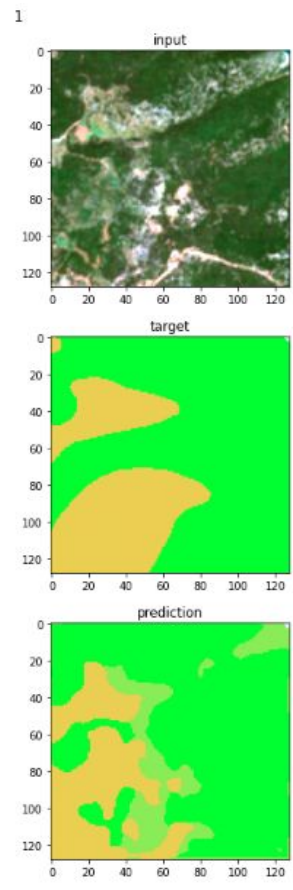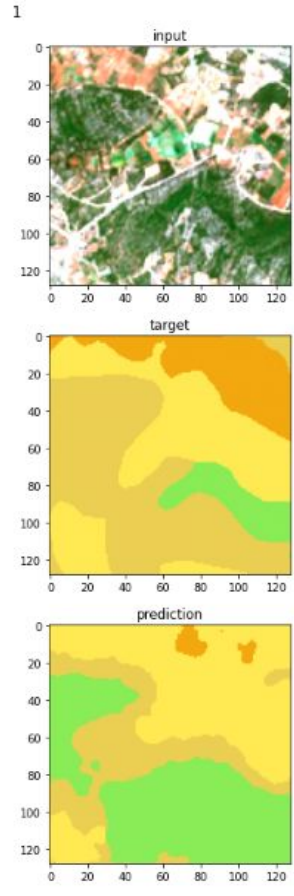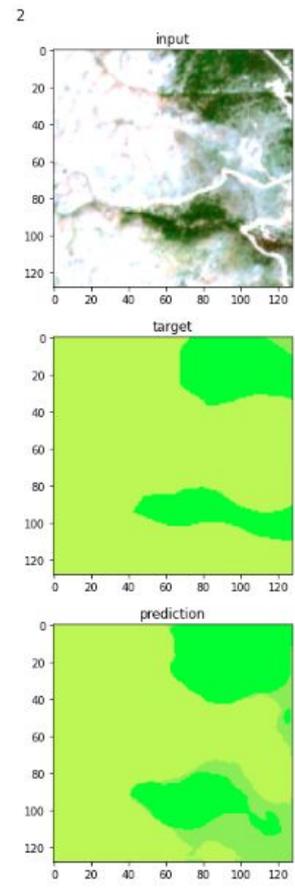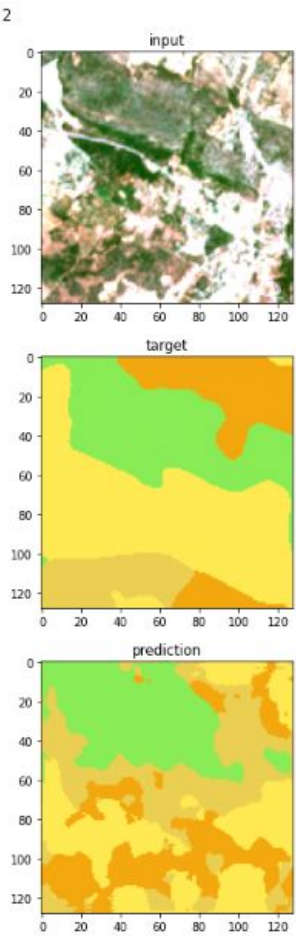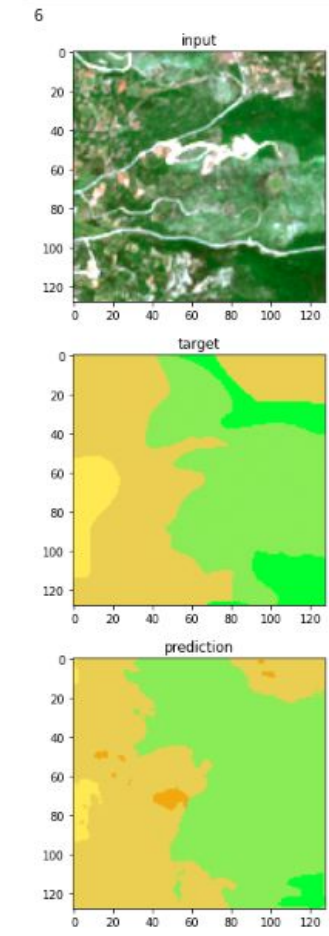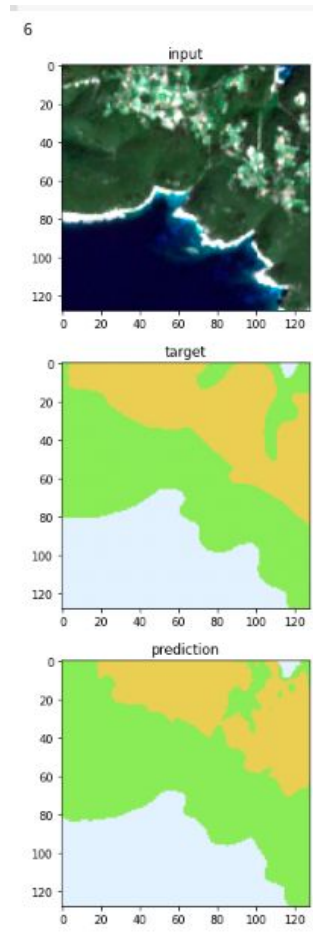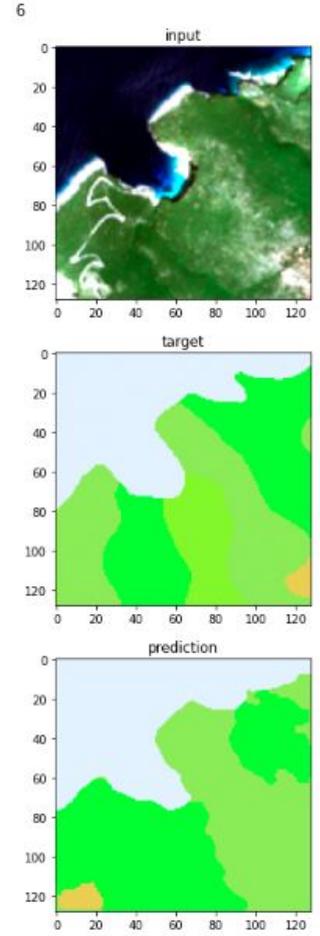

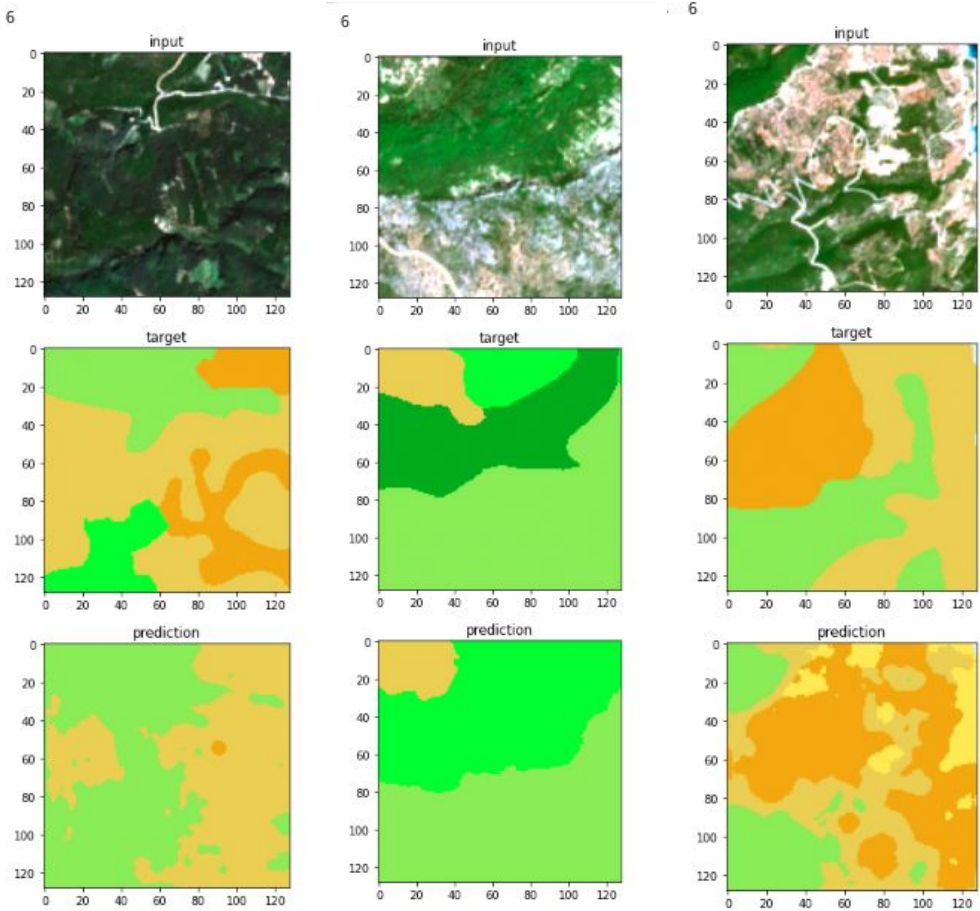

**Figure 16.** Examples of correct prediction of land cover classes/ concord between the predicted and the labeled class boundaries .

In some of the above examples we see the difficulty of our problem, deriving from the low resolution of the input images and the ambiguity of the corine labels. In some cases the model made the right predictions, even though it is a difficult task even for the human observing the rgb input image.

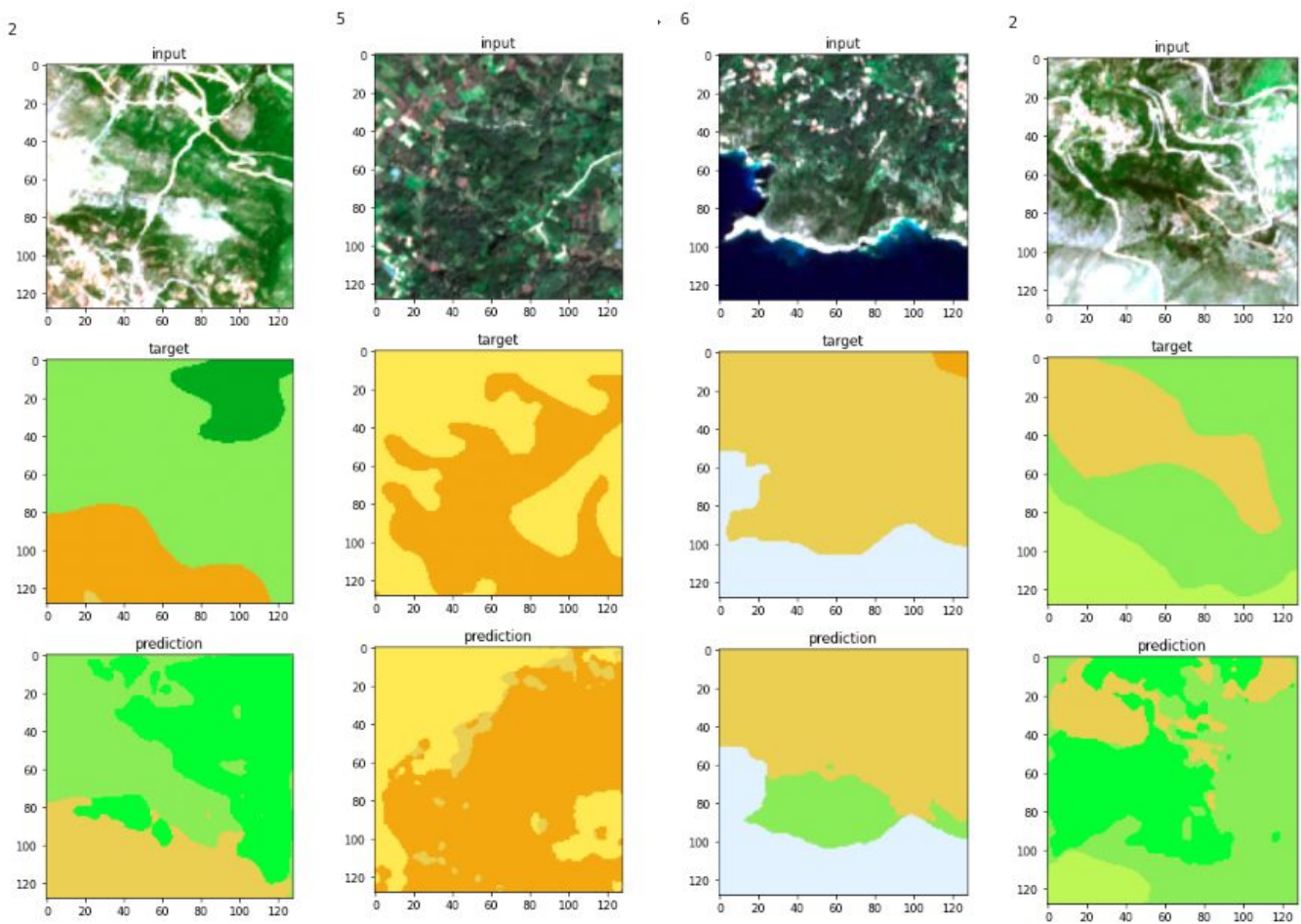

**Figure 17.** Examples that show inaccuracies of the corine land cover that were corrected by our model.

**Conclusion**

- Our models provide a basis for the creation of land cover maps based on the CLC nomenclature. The visual results show the ability of our models to find the boundaries between classes and the accuracy on the higher levels of the class hierarchy is pretty good. The accuracy on common subclasses is also good. However, the performance on predicting uncommon classes and discriminating subclasses of the same superclass on the lower levels of the CLC class hierarchy isn't adequate and human supervision may be needed for this task.
- The CLC dataset contains imperfections. These limit the accuracy of our models. However, in some cases the model can outperform the accuracy of the dataset in cases where the dataset has a lower quality than it's average.
- Usually the land cover is mixed or can not be described accurately by the existing CLC classes. This leads to discord between the labeled data and the predictions, even for kinds of land cover that have been seen on the training set. We also observe that sometimes there are multiple class labels that could describe the land cover and despite the seeming disagreement between the model output and the labels they are close to each other. This indicates the need for a more sophisticated loss criterion and performance metrics that give different penalties to different types of confusion between classes, taking into account the hierarchical structure of the classes and the similarities and overlaps between classes.
- Increasing the resolution of the output from 100m to 10m can give better results but bigger models are required (more parameters).
- The main contribution of transfer learning was speeding up the training processes and possibly improving the results. The encoder part of the network didn't have to be trained, at least for the first epochs of the training, resulting in decreased epoch duration.
- Using a bigger dataset could boost the performance of our models in the area of interest, especially in the task of predicting uncommon classes.

*Key Words: LULC, U-NET, deep learning, transfer learning, Ionio.*